\documentclass[pdflatex,sn-mathphys-num]{sn-jnl}

\usepackage{graphicx}%
\graphicspath{{Figures/}}
\usepackage{multirow}%
\usepackage{amsmath,amssymb,amsfonts}%
\usepackage{amsthm}%
\usepackage{mathrsfs}%
\usepackage[title]{appendix}%
\usepackage{textcomp}%
\usepackage{manyfoot}%
\usepackage{algorithm}%
\usepackage{algorithmicx}%
\usepackage{algpseudocode}%
\usepackage{listings}%
\usepackage{float}          
\usepackage{changepage}     
\usepackage[utf8]{inputenc} 
\usepackage{subcaption} 
\usepackage[nolist]{acronym}
\usepackage{booktabs}                   
\usepackage{multicol}                   
\usepackage{multirow}                   
\usepackage{makecell}
\usepackage[dvipsnames]{xcolor}

\newcommand{\colrect}[2][2.0ex]{%
  \raisebox{-.1\ht\strutbox}{%
    \begingroup
      \setlength{\fboxsep}{0pt}%
      \color{#2}\colorbox{#2}{\rule{#1}{\ht\strutbox}}%
    \endgroup
  }%
}

\definecolor{Background}{RGB}{0,0,0}           
\definecolor{CysticPlate}{RGB}{255,140,0}      
\definecolor{CalotTriangle}{RGB}{255,0,255}    
\definecolor{CysticArtery}{RGB}{220,0,0}       
\definecolor{CysticDuct}{RGB}{0,255,255}       
\definecolor{Gallbladder}{RGB}{0,102,255}      
\definecolor{Tool}{RGB}{170,0,255}             

\definecolor{AbdominalWall}{RGB}{244,172,55}         
\definecolor{Liver}{RGB}{72,160,73}                  
\definecolor{GastrointestinalTract}{RGB}{237,85,59}  
\definecolor{Fat}{RGB}{214,189,153}                  
\definecolor{Grasper}{RGB}{120,90,160}               
\definecolor{LHookElectrocautery}{RGB}{128,128,128}  
\definecolor{Gallbladder2}{RGB}{0,102,255}            
\definecolor{LiverLigament}{RGB}{85,110,200}         

\usepackage{hyperref}       

\theoremstyle{thmstyleone}%

\theoremstyle{thmstyletwo}%

\theoremstyle{thmstylethree}%

\begin{document}

\title[GNNs for Surgical Scene Segmentation]{Graph Neural Networks for Surgical Scene Segmentation}


\author[1,2]{\fnm{Yihan} \sur{Li}}

\author*[1]{\fnm{Nikhil} \sur{Churamani}}\email{nikhil.churamani@medtronic.com}

\author[1]{\fnm{Maria} \sur{Robu}}
\author[1,2]{\fnm{Imanol} \sur{Luengo}}
\author[1,2]{\fnm{Danail} \sur{Stoyanov}}

\affil[1]{\orgname{Medtronic Digital Technologies}, \orgaddress{\city{London}, \country{UK}}}
\affil[2]{\orgname{UCL Hawkes Institute, University College London}, \orgaddress{\city{London}, \country{UK}}}


\abstract{
\textbf{Purpose:} Accurate identification of \textit{hepatocystic} anatomy is critical to preventing surgical complications during \textit{laparoscopic cholecystectomy}. Deep learning models often struggle with occlusions, long-range dependencies, and capturing the fine-scale geometry of rare structures. This work addresses these challenges by introducing \textit{graph-based} segmentation approaches that enhance \textit{spatial} and \textit{semantic} understanding in surgical scene analyses.

\textbf{Methods:} We propose two segmentation models integrating \acf{ViT} feature encoders with \acfp{GNN} to explicitly model spatial relationships between anatomical regions. (1) A \textit{static} \ac{k-NN} graph with a \ac{GCNII} enables stable long-range information propagation. (2) A \textit{dynamic} \ac{DGG} with a \ac{GAT} supports adaptive topology learning. Both models are evaluated on the Endoscapes-Seg50 and CholecSeg8k benchmarks.

\textbf{Results:} The proposed approaches achieve up to $7-8\%$ improvement in \ac{mIoU} and $6\%$ improvement in \ac{mDice} scores over state-of-the-art baselines. It produces anatomically coherent predictions, particularly on thin, rare and safety-critical structures.

\textbf{Conclusion:} The proposed graph-based segmentation methods enhance both performance and anatomical consistency in surgical scene segmentation. By combining \ac{ViT}-based global context with \textit{graph-based} relational reasoning, the models improve interpretability and reliability, paving the way for safer laparoscopic and robot-assisted surgery through a precise identification of critical anatomical features.
}




\maketitle

\section{Introduction}
\label{sec1}

Reliable identification of anatomical structures is critical for surgical safety, particularly in \textit{laparoscopic cholecystectomy}~\cite{ref2}, where major \ac{BDI} often results from misidentifying the cystic duct or artery~\cite{ref3}. The \acf{CVS} protocol~\cite{ref2} was introduced to ensure clear exposure and confirmation of key \textit{hepatocystic} structures before transection, yet achieving \ac{CVS} remains challenging due to occlusions, limited visibility, and visually ambiguous anatomy. These challenges highlight the need for real-time assistive systems that provide robust, anatomically coherent scene understanding. Although recent deep learning–based segmentation methods perform well on anatomical and instrument segmentation, \ac{CNN}- and \acf{ViT}-based models~\cite{ref10,ref13} primarily process local grids or patch sequences without explicitly modelling spatial or semantic relationships, leading to fragmented or topologically inconsistent results, especially for thin or under-represented structures. To address this, we propose a relational reasoning framework using \textit{graph-based learning}, where images are encoded as \textit{graphs} with nodes representing anatomical regions and edges capturing spatial dependencies. \acfp{GNN}~\cite{ref16} thereby enable modelling of both local and global relationships, allowing our proposed methods to learn coherent anatomical structure representations for dense surgical scene segmentation. The main contributions of this work are:
\begin{enumerate}
    \item \textbf{Graph-based Surgical Scene Segmentation:} We apply graph-based learning to model spatial dependencies between anatomical structures for coherent reasoning, proposing two complementary approaches: (i)~a static graph model using \ac{GCNII}~\cite{ref17} for message passing and (ii)~a dynamic method using a \acf{GAT}~\cite{ref16} with \acf{DGG}~\cite{ref19} for adaptive learning.
    \item \textbf{Benchmark Evaluations:} The proposed approaches are evaluated on the Endoscapes-Seg50~\cite{ref3} and CholecSeg8k~\cite{ref20} benchmarks, demonstrating performance gains particularly on thin, safety-critical \textit{hepatocystic} anatomical structures.
\end{enumerate}

\section{Related Work}\label{sec2}
\subsection{Medical Image Segmentation}

Deep learning has revolutionized medical image segmentation through CNN-based architectures such as U-Net~\cite{ref10} and DeepLabV3+~\cite{ref12}, which hierarchically extract local features via convolution and pooling. \acp{ViT}~\cite{ref13} enhance performance by using self-attention to capture global context and hierarchical encoders to process multi-scale features, while foundation models like Med-SAM~\cite{ref25}, trained on large datasets, show strong cross-modality generalization. Anatomy segmentation~\cite{ref2,ref3}, however, still remains a complex task, requiring delineation of deformable, texture-similar, and occluded structures with topological fidelity. Most approaches employ deep decoders with multi-scale fusion or domain adaptation to address inter-patient variability, yet they still operate at the pixel level without enforcing relational consistency—motivating representations that model anatomy as connected, interdependent regions rather than independent pixels.

\subsection{Graph-Based Learning}
Graph learning on structured image data represents an image as a graph $G = (V, E)$, where $V$ is the set of nodes representing image regions 
and $E$ denotes the edges encoding spatial adjacency or feature similarity between nodes. The goal is to learn a function $f:(X, A) \mapsto Y$, where $X$ is the node feature matrix, $A$ is the adjacency structure, and $Y$ are node level predictions.



\subsubsection{Node Construction and Feature Extraction}

Graph nodes represent coherent image regions, such as superpixels, which align with object boundaries and reduce graph size while preserving contours. Some approaches instead construct graphs over fixed grid patches, consistent with \ac{ViT} tokenization, where each patch serves as a node with associated feature embeddings~\cite{ref13}. In video settings, nodes can persist across frames to form spatio-temporal graphs~\cite{ref34} that maintain identity under motion and occlusion. Each node descriptor ($\mathbf{x}(v_i)$) encodes visual and contextual properties. While hand-crafted features (e.g., mean color, centroid, texture) are efficient but limited, recent methods pool deep encoder features to obtain semantically rich embeddings. Self-supervised \ac{ViT} embeddings learned via masked reconstruction~\cite{ref37} or distillation~\cite{ref33} further enhance representation quality by clustering semantically coherent regions and preserving geometric relationships.

\subsubsection{Edge Construction}
Edges determine the propagation of information in the graph, encoding inductive bias. Region Adjacency Graphs~\cite{ref30} connect spatial neighbours, preserve topology and support boundary-sensitive propagation, but are local. 
In contrast, feature-affinity graphs can connect $k$ nearest neighbours in the embedding space, enabling distant yet semantically related connections. Furthermore, dynamic graphs~\cite{ref34} jointly optimise connectivity with the downstream network, blending geometric locality with semantic similarity. In surgical scenes, temporal edges~\cite{ref34} linking regions across frames can further enhance coherence under occlusion and motion.


\subsection{Graph Neural Networks}

\acp{GNN}~\cite{ref16,ref17,ref19} operate through \textit{message passing} and \textit{aggregation}, where nodes exchange and integrate information from their neighbours. Among key variants, \ac{GCNII}~\cite{ref17} enhances depth stability for long-range propagation, while \ac{GAT}~\cite{ref16} introduces attention-based, content-adaptive aggregation in heterogeneous neighbourhoods. In surgical scene analysis, extending \acp{GNN} from dynamic object-level~\cite{ref34} to region-level (patch or superpixel) graphs enables continuity along thin structures and focused inference on critical, visually ambiguous interfaces.

\section{Methods}
\label{sec3}

\begin{figure}[t]
    \centering
    \includegraphics[width=1\linewidth]{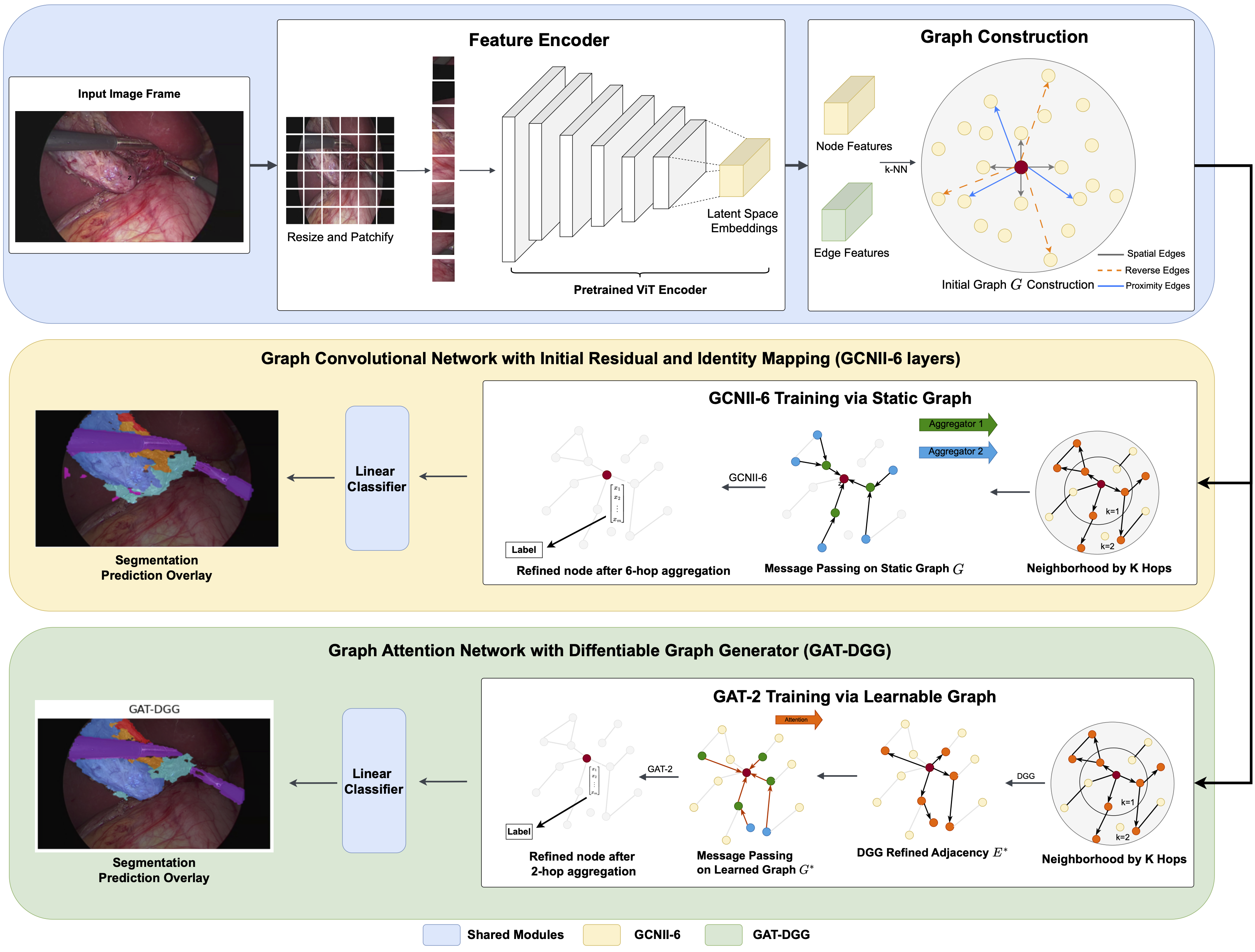}
    \caption{\textbf{Proposed \ac{GCNII}-6} and \textbf{\ac{GAT}-\ac{DGG}} use \textbf{\ac{ViT}}-based encoders to construct a graph (for each dataset) offline. \ac{GCNII}-6 uses a static graph while \ac{GAT}-\ac{DGG} dynamically adapts the topology of this graph, both learning to segment different structures.}
    \label{fig:overview}
\end{figure}
Fig.~\ref{fig:overview} provides an overview of the proposed graph-based segmentation approaches. Both methods share feature extraction and graph construction modules. Each frame $I_t$ passes through four main modules: \textbf{(1) Feature extraction:} a pre-trained encoder generates feature embeddings, partitioned into patches that serve as graph nodes; \textbf{(2) Graph construction:} a graph $\mathcal{G}_t = (\mathcal{V}_t, \mathcal{E}_t)$ is constructed offline, with each node representing local patch features and edges connecting the nodes; \textbf{(3) \acp{GNN} training:} two alternative end-to-end models are proposed. Firstly, \textbf{\ac{GCNII}-6} operates on a static graph, while the second employs \textbf{\ac{GAT} with \ac{DGG}}, where adjacency is refined online in a task-dependent manner; \textbf{(4) Segmentation output:} a linear upsampler restores full-resolution masks.


\subsection{Feature Encoder}
\label{sec:encoder}
Input frame $I_t$ is partitioned into $N, P \times P$ patches, each flattened and linearly projected into $D$-dimensional token embeddings.
These tokens are passed to the pre-trained \ac{ViT} encoder to capture long-range features, invariant to intraoperative noise. We use two \textit{pre-trained} and \textit{frozen} encoders in our evaluations. Firstly, EndoViT~\cite{ref50} is used which is a domain-adapted \ac{ViT}-\ac{MAE}~\cite{ref37} pre-trained on a large corpus of laparoscopic and robotic frames, robustly encoding global shape and boundary structures. The output embeddings provide mid- to high-level features robust to specularities, occlusion, and deformation, conditions ubiquitous in endoscopic images. Secondly, we also use \ac{ViT}-DINO~\cite{ref33} trained by self-distillation without labels. It aligns a student network output distribution with that of a momentum-updated teacher via a cross-entropy loss. This enforces invariance across spatial augmentations and illumination changes, yielding semantically structured representations complementary to the geometry-aware embeddings from \ac{MAE}. We use the pre-trained weights provided for both EndoViT~\cite{ref50} and \ac{ViT}-DINO~\cite{ref33} encoders.


\subsection{Graph Construction}
From patch embeddings, a sparse weighted graph $G = (V, E)$ is constructed over the $N$ patches. Each node $v_i \in V$ represents a feature-vector $\mathbf{x}_i = \mathbf{h}_i \in \mathbb{R}^D$ with patch centre $p_i$. Edges combine:~(i) spatial adjacency to preserve topology and~(ii) feature-affinity \ac{k-NN} to connect semantically related albeit distant regions. To improve boundary calibration, we propose to include a small number of contrastive \textit{heterophilous} links to dissimilar patches, which are down-weighted to avoid destabilizing propagation. For $i\!\to\! j$ we assign a Gaussian kernel weight $w_{ij}\ \propto\ \gamma_{ij}\,\exp\!\Big(
-\frac{(1-\mathbf{x}_i^\top\mathbf{x}_j)^2}{2\sigma_f^2}
-\frac{\lVert \mathbf{p}_i-\mathbf{p}_j\rVert_2^2}{2\sigma_s^2}
\Big)$ and row-normalize to obtain a directed, degree-normalized adjacency $A$.


\subsection{\acf{GNN} Training}
\ac{GNN} layers update nodes via message passing and permutation-invariant aggregation:
\begin{equation}
\mathbf{h}_i^{(l+1)} = \psi \!\left( 
\mathbf{h}_i^{(l)}, 
\bigoplus_{j \in \mathcal{N}(i)} 
\phi(\mathbf{h}_i^{(l)}, \mathbf{h}_j^{(l)}, A_{ij})
\right)
\end{equation}
where $\phi$ defines how messages are constructed, $\bigoplus$ denotes the aggregation function, and $\psi$ updates the node embeddings. By stacking such layers, \acp{GNN} incorporate increasingly long-range relational context. The classical \acf{GCN}~\cite{ref17} can be viewed as a special case in which messages are averaged through the normalized adjacency matrix $\tilde{A} = D^{-\frac{1}{2}} (A + I) D^{-\frac{1}{2}}$.
Standard \acp{GCN} suffer from over-smoothing, where node embeddings become indistinguishable as depth increases, limiting their ability to model multi-hop dependencies.
To enable deeper, more stable propagation, we propose the use of  \textbf{\ac{GCNII}}~\cite{ref17} with $6$ layers, operating on the pre-constructed graph, where a layer is updated as:
\begin{equation}
\mathbf{H}^{(l+1)} = 
\sigma \!\big(
A \big[(1 - \alpha_l)\mathbf{H}^{(l)} + \alpha_l \mathbf{H}^{(0)}\big] \mathbf{W}_l
\big)    
\end{equation}
where  $\mathbf{W}_l$ are trainable weights and $\alpha_l$ controls the influence of the initial embeddings $\mathbf{H}^{(0)}$. This prevents feature collapse and allows propagation over $6+$ layers without degradation.  

Alternatively, instead of using a fixed graph, we propose the use of \textbf{\ac{GAT}}~\cite{ref16} \textbf{with \ac{DGG}}~\cite{ref19} to dynamically adapt the graph topology where instead of fixed weights, each node computes content-dependent attention coefficients with its neighbours:
\begin{equation}
e_{ij} = \mathrm{LeakyReLU}\big(\mathbf{a}^{\top}[\mathbf{W}\mathbf{h}_i \,\|\, \mathbf{W}\mathbf{h}_j]\big),
\quad
A_{ij}^{\mathrm{dyn}} = \frac{\exp(e_{ij})}{\sum_{k \in \mathcal{N}(i)} \exp(e_{ik})}
\end{equation}
The \ac{DGG} further refines this learned adjacency by gating edge strengths:
\begin{equation}
\tilde{A}_{ij} = A_{ij}^{\mathrm{dyn}} \cdot \sigma(g_{ij}), 
\quad 
g_{ij} = f_{\mathrm{MLP}}([\mathbf{h}_i, \mathbf{h}_j])
\end{equation}
allowing edge activations to be optimized jointly with node embeddings. 

\subsection{Segmentation Decoder and Loss Function}
Finally, node embeddings are projected back to the pixel domain for dense mask prediction. Given node logits $\mathbf{Z} \in \mathbb{R}^{N \times C}$, class posteriors are obtained by softmax normalisation and reshaped to the encoder's token grid $\Pi_{\mathrm{grid}} \in \mathbb{R}^{H_s \times W_s \times C}$ . We recover full-resolution posteriors via a parameter-free bilinear projection that linearly interpolates class probabilities. 



\noindent
\textbf{Losses:} 
We combine complementary objectives: 
\begin{equation}
\mathcal{L} = 
\lambda_{\mathrm{CE}} .  \mathcal{L}_{\mathrm{CE}} 
+ \lambda_{\mathrm{Dice}} . \mathcal{L}_{\mathrm{Dice}} 
+ \lambda_{\mathrm{Lov}} . \mathcal{L}_{\mathrm{Lov\acute{a}sz}} 
+ \lambda_{\mathrm{Potts}} . \mathcal{L}_{\mathrm{Potts}}
\label{eq:loss}
\end{equation}
where, class-balanced cross-entropy 
stabilizes gradients under long-tailed distributions
Dice loss improves small-region recall, Lovász–Softmax~\cite{ref55} aligns optimization with \ac{IoU} evaluations by ranking dominant pixel errors. Finally, boundary-aware Potts regulariser enforces spatial smoothness.
Such a composite loss provides stable optimization under strong imbalances and preserves boundaries.

\section{Experiments}\label{sec4}
\textbf{Dataset:} The models are evaluated on two public laparoscopic cholecystectomy benchmarks: Endoscapes–Seg$50$~\cite{ref3} comprising of $493$ frames, across $50$ videos, with pixel-wise masks for $6$ CVS-relevant classes including \textit{cystic plate, HC-triangle, cystic artery, cystic duct, gallbladder} and \textit{tools}. A \verb|train:val:test| split of $30$:$10$:$10$ is used on videos. CholecSeg8k~\cite{ref20} contains $8080$ annotated frames from $17$ procedures covering $13$ classes: \textit{background, abdominal wall, liver, gastrointestinal tract, fat, grasper, connective tissue, blood, cystic duct, L-hook electrocautery, gallbladder, hepatic vein}, and \textit{liver ligament}. We follow the video split proposed in~\cite{ref56} with $75\%$ videos are used for training and $25\%$ for validation (videos $12, 20, 48$ and $55$).

\noindent
\textbf{Metrics:} Macro-averaged \acf{mIoU} and \acf{mDice} scores are reported for each class. 

\noindent
\textbf{Implementation:} Frames are resized to $1024 \times 1024$, encoded using the \textit{frozen} \ac{ViT} backbones to stride-4 tokens ($128 \times 128$ grid; $N = 16,384$ nodes). EndoViT is used for Endoscapes-Seg50 while \ac{ViT}-DINO is used for CholecSeg8k to avoid data leakage as EndoViT was pretrained with CholecSeg8k in the corpus. Optimization uses AdamW with mixed precision, over $100 epochs$. On Endoscapes–Seg50, we train with cosine decay and a base \acs{LR} $=2e^{-5}$. On CholecSeg8k, we use \textit{One-CycleLR} ($pct_{start}=0.3$) and a base \acs{LR} $=5e^{-5}$. All results are reported from the best-val checkpoint.

\noindent
\textbf{Loss Coefficients:} On cleaner and reletively balanced Endoscapes-Seg50, a calibrated \textit{class-weighed cross-entropy} loss provides the best results. CholecSeg8k exhibits higher class imbalances and noisier boundaries; thus a composite objective is used as described in Eq.~\ref{eq:loss} with $\lambda_{\mathrm{CE(cb-\sqrt{\cdot})}}=0.6, \lambda_{\mathrm{Dice}}=0.2, \lambda_{\mathrm{Lov}}=0.2$ and $\lambda_{\mathrm{Potts}}=0.05$.

\section{Results and Discussion}
\label{sec5}
\subsection{Results}

\begin{table}[t]
\centering
\caption{\acs{mIoU} and \acs{mDice} Scores on Endoscapes-Seg50.}
\begin{tabular}{l c c}\toprule
       \textbf{Model} & \textbf{\acs{mIoU}} & \textbf{\acs{mDice}} \\ \midrule

DLV3P-R50-SurgMoCov2~\cite{ref61} & 0.1570 & 0.2405 \\

CycleSAM-DLV3P-SurgMoCov2 \cite{ref57, ref61}& 0.2234 & 0.3086 \\ 

DLV3P-R50-FullSup \cite{ref12}& 0.4502 & 0.5760 \\
EndoViT+DPT Baseline~\cite{ref50} & 0.4604 & 0.5942 \\ \midrule 

 \textbf{GCNII-6 (Ours)} & \textbf{0.5358}  & \textbf{0.6553} \\
 \textbf{GAT-DGG (Ours) }  & \textbf{0.5384} & \textbf{0.6572}\\ \bottomrule
\end{tabular}
\label{tab:es50_results}
\end{table}

\begin{figure}[t]
    \centering
    \includegraphics[width=\linewidth]{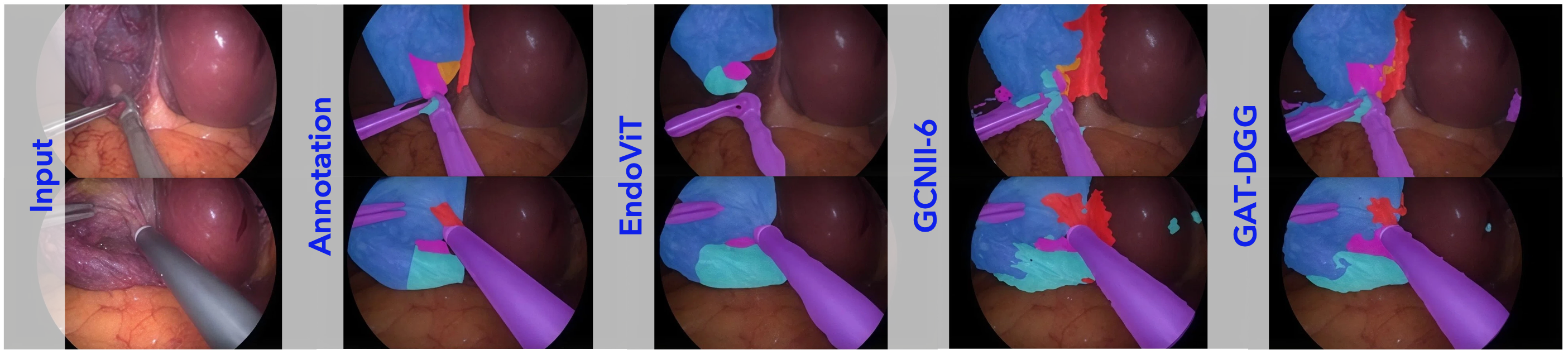}
    \caption{\textbf{Qualitative Results on Endoscapes-Seg50.} Classes:
    background~\colrect{Background}, 
    cystic~plate~\colrect{CysticPlate}, 
    calot’s~triangle~\colrect{CalotTriangle}, 
    cystic~artery~\colrect{CysticArtery}, 
    cystic~duct~\colrect{CysticDuct}, 
    gallbladder~\colrect{Gallbladder}, 
    tool~\colrect{Tool}.}
    \label{fig:endo_prediction}
\end{figure}

\begin{table}[b]
\centering
\caption{Per-class IoU (left) and Dice (right) scores on Endoscapes-Seg50.}
\begin{tabular}{lccc|ccc}\toprule
& \multicolumn{3}{c|}{\textbf{\acs{IoU} Scores}} & \multicolumn{3}{c}{\textbf{Dice Scores}} \\ \cmidrule{2-7}

\textbf{Class}            &\makecell{\textbf{EndoViT} \\ \textbf{Baseline}} & \makecell{\textbf{GCNII-6} \\ \textbf{(Ours)}} & \makecell{\textbf{GAT-DGG} \\ \textbf{(Ours)}} &\makecell{\textbf{EndoViT} \\ \textbf{Baseline}}  & \makecell{\textbf{GCNII-6} \\ \textbf{(Ours)}} & \makecell{\textbf{GAT-DGG} \\ \textbf{(Ours)}} \\\midrule

Background       &\textbf{0.9474} & 0.9305 & 0.9280 &\textbf{0.9673}  & 0.9640 & 0.9605      \\
Cystic Plate     & 0.2497 & 0.4409 & \textbf{0.4512} & 0.3996 & 0.6120 & \textbf{0.6180}    \\
HC Triangle & 0.0879 & 0.1883 & \textbf{0.2045} & 0.1616 & 0.3169 & \textbf{0.3319} \\
Cystic Artery    & 0.1197 & \textbf{0.3085} & 0.2997  & 0.2138 & \textbf{0.4715} & 0.4544 \\
Cystic Duct      & 0.2981 & 0.3390 & \textbf{0.3542} & 0.4593 & 0.5063 & \textbf{0.5174}    \\
Gallbladder      & \textbf{0.8082} & 0.7822 & 0.7755 & \textbf{0.8939} & 0.8778 & 0.8571    \\
Tool             & 0.7452 & 0.7993 & \textbf{0.8170} & 0.8540 & 0.8885 & \textbf{0.8923}     \\\midrule
Mean             & 0.4604 & 0.5358 & \textbf{0.5384} & 0.5942 & 0.6553 & \textbf{0.6572}       \\
\bottomrule
\end{tabular}
\label{tab:endo_iou_results}
\end{table}
Table~\ref{tab:es50_results} summarizes results on Endoscapes-Seg50, showing that both \ac{GCNII}-6 and \ac{GAT}-\ac{DGG} substantially outperform state-of-the-art baselines, achieving $+7-8\%$ \ac{mIoU} and $+6\%$ \ac{mDice} scores. These gains demonstrate that graph reasoning provides consistent improvements beyond \ac{ViT}-based encoder features. Per-class results (Table~\ref{tab:endo_iou_results}) reveal that improvements are most pronounced for thin, rare structures such as the cystic artery, cystic duct, cystic plate, and HC-triangle where baselines often fragment or blur boundaries, while larger, texture-rich classes like the gallbladder remain comparable. The performance boost arises from integrating local geometric and global semantic features propagation through feature-based \ac{k-NN}, spatial, and sparse reverse edges. \ac{GCNII}-6 achieves stable high-order diffusion via its residual formulation, reinforcing thin-structure continuity without over-smoothing, whereas \ac{GAT}-\ac{DGG} dynamically refines edge importance through attention. Both models enhance boundary coherence and continuity, reconnecting occluded anatomy traces that EndoViT tends to truncate (Fig.~\ref{fig:endo_prediction}). \ac{GAT}-\ac{DGG} generally yields smoother results, while \ac{GCNII}-6 may slightly over-segment yet remains topologically consistent.

\begin{table}[b]
\centering
\caption{Per-class IoU scores for Anatomy Classes on CholecSeg8k.}
\label{tab:cholec_results}
\begin{tabular}{lccc|cc} 
\toprule
\textbf{Class} & \makecell{\textbf{Mask2Former} \\ \cite{ref56}}  & \makecell{\textbf{Swin Base} \\ \cite{ref56}}  & \makecell{\textbf{SP-TCN} \\ \cite{ref56}}  & \makecell{\textbf{GCNII-6} \\ (Ours)} & \makecell{\textbf{GAT-DGG} \\ (Ours)} \\
\midrule
Background          &   0.9785  & 0.9731 & 0.9740  & 0.9713 & \textbf{0.9847} \\
Abdominal wall      &   0.6925  & 0.7848 & 0.7414  & 0.7316 & \textbf{0.7991} \\
Connective tissue   &   0.2484  & 0.2510 & 0.3120  & 0.4990 & \textbf{0.5332} \\
Fat                 & \textbf{0.8417} & 0.8408  & 0.8415 & 0.8048 & 0.8067 \\
Gallbladder         & \textbf{0.6520} & 0.6040  & 0.6109  & 0.5679 & 0.5794 \\
Gastrointestinal~tract & 0.5230 & 0.4935 & 0.5714  & 0.5542 & \textbf{0.5808} \\
Liver               & 0.8000 & \textbf{0.8016} & 0.7750  & 0.7529 & 0.7503 \\ \midrule
Mean                & 0.6766 & 0.6784 & 0.6895 & 0.6974 & \textbf{0.7192} \\ \bottomrule
\end{tabular}

\footnotetext[1]{Tool IoU Results: \\
\textbf{Grasper}: Swin Base: 0.7263 $|$ GCNII-6: 0.6265 $|$ GAT-DGG: 0.6461 \\
\textbf{L-Hook}~: Swin Base: 0.6829~$|$~GCNII-6: 0.5996~$|$~GAT-DGG: 0.6247
}
\end{table}

\begin{figure}[t]
    \centering
    \includegraphics[width=\linewidth]{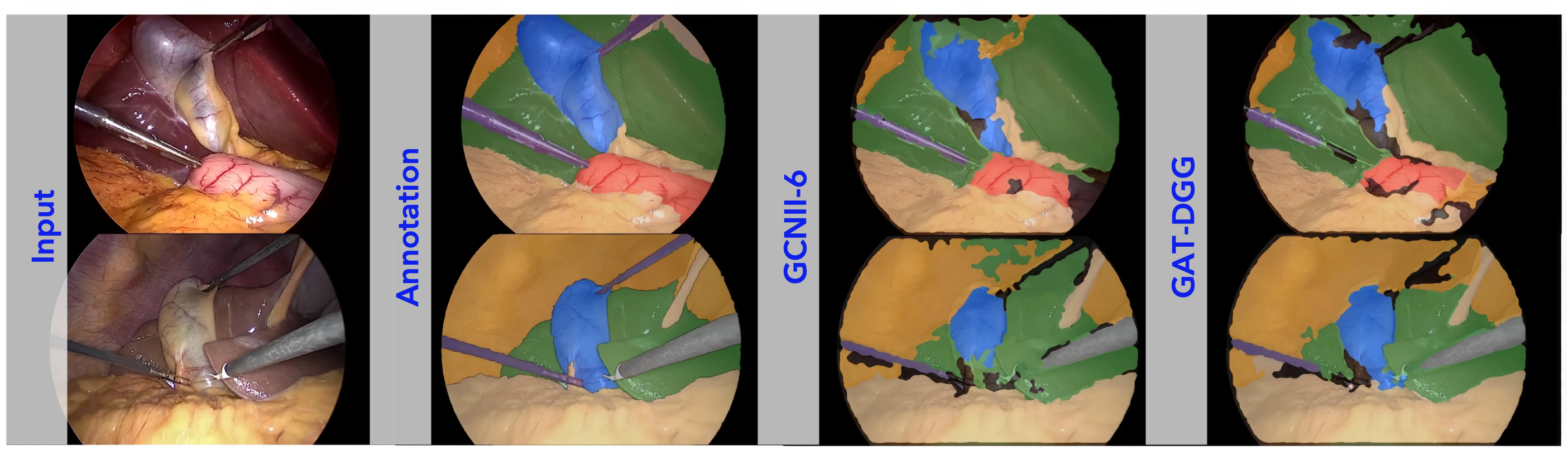}
    \caption{\textbf{Qualitative Results on CholecSeg8k.} Classes: background~\colrect{Background}, 
    abdominal~wall~\colrect{AbdominalWall}, 
    liver~\colrect{Liver}, 
    gastrointestinal~tract~\colrect{GastrointestinalTract}, 
    fat~\colrect{Fat}, 
    grasper~\colrect{Grasper}, 
    L-hook~electrocautery~\colrect{LHookElectrocautery}, 
    gallbladder~\colrect{Gallbladder2}, 
    liver~ligament~\colrect{LiverLigament}.}
    \label{fig:cholec_prediction}
\end{figure}

On CholecSeg8k (Table~\ref{tab:cholec_results}), we report per-class and mean \ac{IoU} scores only for anatomical structures, as graph-based methods focus on modelling spatial and semantic relationships to enforce anatomical continuity. The evaluation uses a \textit{pre-trained} \ac{ViT}-DINO encoder to avoid data leakage from EndoViT \textit{pre-training}. Both \ac{GCNII}-6 and \ac{GAT}-\ac{DGG} outperform baselines, with \ac{GAT}-\ac{DGG} improving $2\text{--}3\%$ over the best SP-TCN~\cite{ref56}, which encodes spatio-temporal features. Similar to EndoScapes-Seg50, gains are largest on thin, boundary-sensitive classes like connective tissue, while large homogeneous organs show smaller improvements, highlighting the benefit of structured graph reasoning for rare or thin structures even without temporal modelling. Qualitative results (Fig.~\ref{fig:cholec_prediction}) show coherent anatomical boundaries across varying camera poses and lighting. \ac{GCNII}-6 captures global topology of large tissues but may miss thin structures, whereas \ac{GAT}-\ac{DGG} better delineates fine, high-curvature structures and tool-tissue junctions via dynamic edge weighting, preserving overall structural consistency.



\subsection{Ablations}


\noindent
\textbf{Encoder:} We compare EndoViT~\cite{ref50}, \ac{ViT}-DINO~\cite{ref33}, \ac{ViT}-\ac{MAE}~\cite{ref37}, and MedSAM~\cite{ref25} using k-means clustering and downstream segmentation. EndoViT produces the most anatomy-aligned clusters with the best accuracy–cost trade-off, while \ac{ViT}-DINO yields finer granularity aiding thin structures but over-segments textures and quadruples node/edge counts, increasing memory and compute costs.


\noindent
\textbf{Graph Construction:} Using EndoViT, we evaluate connectivity with \ac{GCNII}-6. A hybrid graph with $8$ \ac{k-NN} feature edges, $8$ spatial edges, and $4$ farthest reverse edges performs the best. Spatial edges stabilize training, \ac{k-NN} links to distant \textit{homologous} regions, and down-weighted reverse edges introduce controlled \textit{heterophily}, improving boundary calibration.

\noindent
\textbf{\acp{GNN}:} On the hybrid graph, \ac{GCNII}-6 outperforms \ac{GCN}, GraphSAGE~\cite{ref44}, \ac{GAT}~\cite{ref16}, and Graph Transformer~\cite{ref45} via depth-stable diffusion mitigating over-smoothing. On \ac{DGG}-learned graphs, \ac{GAT} excels by attention-based edge selection. 

Together, \textbf{encoder features} define semantic neighbourhoods, \textbf{hybrid graphs} encode geometric and \textbf{heterophilous} priors, and \textbf{depth-stable propagation} converts local cues into \textbf{globally coherent, boundary-sensitive masks}, benefiting CVS-critical structures.


\subsection{Graph Reasoning}
\begin{figure}[t]
    \centering
    \includegraphics[width=1\linewidth]{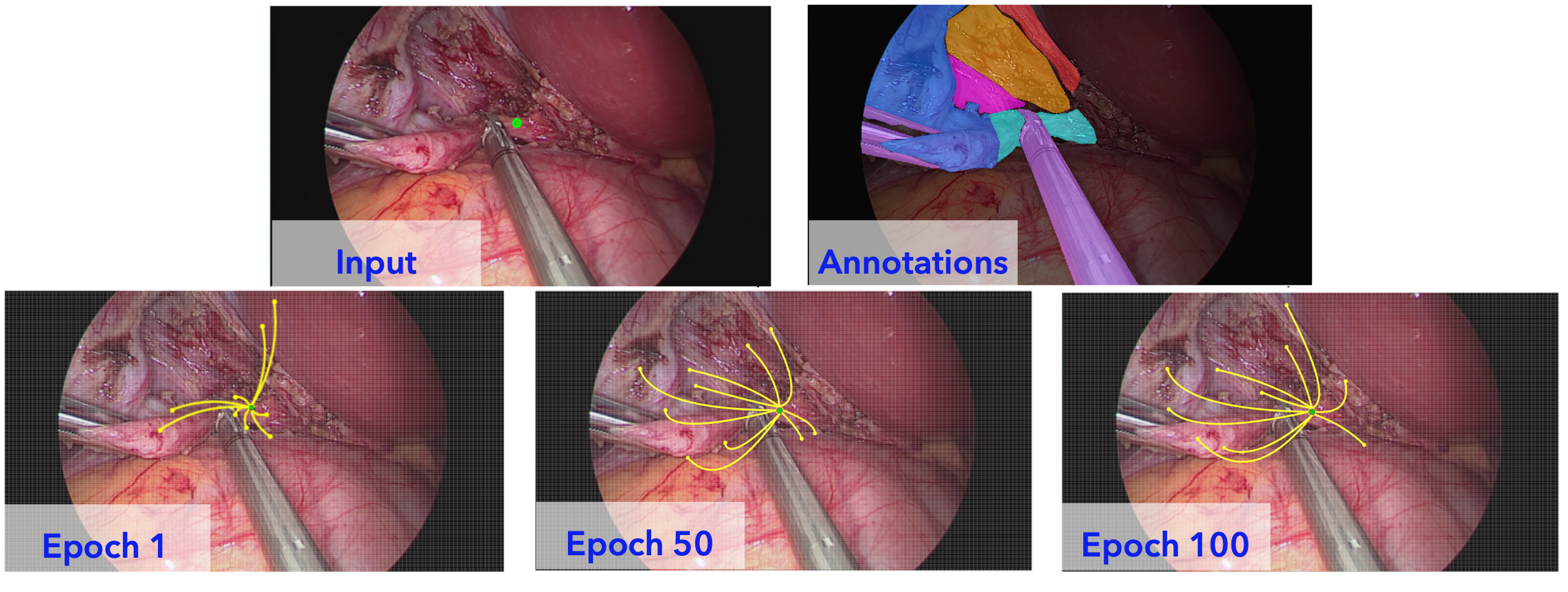}
        \caption{Visualisation on Endoscapes-Seg50 of a \textit{cystic duct} node learning long-range dependencies across neighbourhood nodes. Classes:
    background~\colrect{Background}, 
    cystic~plate~\colrect{CysticPlate}, 
    calot’s~triangle~\colrect{CalotTriangle}, 
    cystic~artery~\colrect{CysticArtery}, 
    cystic~duct~\colrect{CysticDuct}, 
    gallbladder~\colrect{Gallbladder}, 
    tool~\colrect{Tool}.}
    \label{fig:reasoning}
\end{figure}


Fig.~\ref{fig:reasoning} illustrates how graph message passing captures long-range dependencies for the \textit{cystic duct}, a thin and variable structure difficult to segment using local appearance alone. Early in training (Epoch~$1$), the target node connects mainly to short, local edges along the duct and occasionally to instruments, reflecting reliance on local texture and high-contrast edges. By Epoch~$50$, connections extend toward the \textit{gallbladder neck, Calot’s triangle}, and the \textit{cystic artery}, indicating the model begins to leverage surrounding anatomical context. By Epoch $100$, the strongest edges consistently link the duct to these regions, while connections to tools and background weaken. This demonstrates that cystic duct is recognized not in isolation but via its relational position within \textit{hepatocystic} anatomy, reflecting successful learning of long-range dependencies.

\section{Conclusion}\label{sec6}

This work presents two graph-based segmentation approaches that integrate \ac{ViT} features with \acp{GNN} for anatomically consistent surgical scene understanding. Each frame is processed as a graph of patch embeddings connected via spatial and semantic affinities, capturing local boundaries and global context to improve upon deep-learning baselines across two benchmarks. \ac{GCNII}-6 uses a static graph for robust generalization under noise and class imbalance, while \ac{GAT}-\ac{DGG} dynamically learns graph topology, refining context via attention. Both demonstrate that explicit relational reasoning benefits rare, safety-critical anatomies. 
Future work will extend these methods to spatio-temporal graphs~\cite{ref34} to enforce temporal consistency, explore hierarchical or multi-scale graphs~\cite{ref63} to enrich node and edge features, and optimize for real-time performance to bridge the gap between benchmarks and surgical deployment.

\backmatter

\section*{Statements and Declarations}

\begin{itemize}
    \item \textbf{Competing Interests:} NC, MR, IL and DS are employees of Medtronic Digital Technologies. YL contributed to this work as part of their MSc dissertation conducted as a Research Intern at Medtronic Digital Technologies.
\end{itemize}

\begin{acronym}
    \acro{BDI}{Bile Duct Injury}
    \acro{CNN}{Convolutional Neural Network}
    \acro{CVS}{Critical View of Safety}
    \acro{DGG}{Differentiable Graph Generator}
    \acro{k-NN}{k Nearest Neighbours}
    \acro{ViT}{Vision Transformer}
    \acro{GAT}{Graph Attention Network}
    \acro{GCN}{Graph Convolutional Network}
    \acro{GCNII}{Graph Convolutional Network with Initial Residual and Identity Mapping}
    \acro{GNN}{Graph Neural Network}
    \acro{IoU}{Intersection over Union}
    \acro{LR}{Learning Rate}
    \acro{MAE}{Masked Autoencoder}
    \acro{mIoU}{Mean Intersection over Union}
    \acro{mDice}{Mean Dice}
    \acro{SAM}{Segment Anything Model}
    \acro{SLIC}{Simple Linear Iterative Clustering}
\end{acronym}

\bibliography{sn-bibliography}

\end{document}